\documentclass[final]{cvpr}

\usepackage{times}
\usepackage{epsfig}
\usepackage{graphicx}
\usepackage{amsmath}
\usepackage{amssymb}
\usepackage{multirow}
\usepackage{enumerate}
\usepackage{color,xcolor}
\usepackage[switch]{lineno}
\usepackage{threeparttable}
\DeclareMathOperator*{\argmax}{argmax}

\usepackage[pagebackref=true,breaklinks=true,colorlinks,bookmarks=false]{hyperref}

\def\cvprPaperID{3575} 
\def\confYear{CVPR 2021}

\begin{document}

\title{Self-Guided and Cross-Guided Learning for Few-Shot Segmentation}

\author{Bingfeng Zhang\textsuperscript{1,2},~~~~~~~
	Jimin Xiao\textsuperscript{1}\thanks{Corresponding author}~~,~~~~
	Terry Qin\textsuperscript{3}\\
{\textsuperscript{1}XJTLU,~~~~}
{\textsuperscript{2}University of Liverpool,~~~~}
{\textsuperscript{3} Dinnar Automation Technology}\\
{\tt\small \{Bingfeng.Zhang,Jimin.Xiao\}@xjtlu.edu.cn},~~ {\tt\small terry.qin@outlook.com}\\
}

\maketitle
\pagestyle{empty}
\thispagestyle{empty}

\begin{abstract}
   Few-shot segmentation has been attracting a lot of attention due to its effectiveness to segment unseen object classes with a few annotated samples. Most existing approaches use masked Global Average Pooling (GAP) to encode an annotated support image to a feature vector to facilitate query image segmentation.
   However, this pipeline unavoidably loses some discriminative information due to the average operation. 
   In this paper, we propose a simple but effective self-guided learning approach, where the lost critical information is mined. Specifically, through making an initial prediction for the annotated support image, the covered and uncovered foreground regions are encoded to the primary and auxiliary support vectors using masked GAP, respectively. By aggregating both primary and auxiliary support vectors, better segmentation performances are obtained on query images. 
   Enlightened by our self-guided module for 1-shot segmentation, we propose a cross-guided module for multiple shot segmentation, where the final mask is fused using predictions from multiple annotated samples with high-quality support vectors contributing more and vice versa. 
   This module improves the final prediction in the inference stage without re-training.  
   Extensive experiments show that our approach achieves new state-of-the-art performances on both PASCAL-$\text{5}^i$ and COCO-$\text{20}^i$ datasets. Source code is available at \href{https://github.com/zbf1991/SCL}{https://github.com/zbf1991/SCL}
   \footnotetext[1]{The work was supported by National Natural Science Foundation of China under 61972323.}.
\end{abstract}

\section{Introduction}
Semantic segmentation has been making great progress with recent advances in deep neural network especially Fully Convolutional Network (FCN)~\cite{long2015fully}. Requiring sufficient and accurate pixel-level annotated data, state-of-the-art semantic segmentation approaches can produce satisfying segmentation masks. However, these approaches heavily rely on massive annotated data.
Their performance drops dramatically on unseen classes or with insufficient annotated data~\cite{zhang2020reliability}.

Few-shot segmentation~\cite{gairola2020simpropnet, li2020fss, rakelly2018conditional, siam2019amp} is a promising method to tackle this issue. Compared to fully supervised semantic segmentation~~\cite{chen2018deeplab, chen2018encoder, Huang_2019_ICCV, li2020improving} which can solely segment the same classes in the training set, the objective of few-shot segmentation is to utilize one or a few annotated samples to segment new classes. Specifically, the data in few-shot segmentation is divided into two sets: support set and query set. This task requires to segment images from the query set given one or several annotated images from the support set. Thus, the key challenge of this task is how to leverage the information from the support set.
\begin{figure}
	\centering
	\includegraphics[width=\columnwidth]{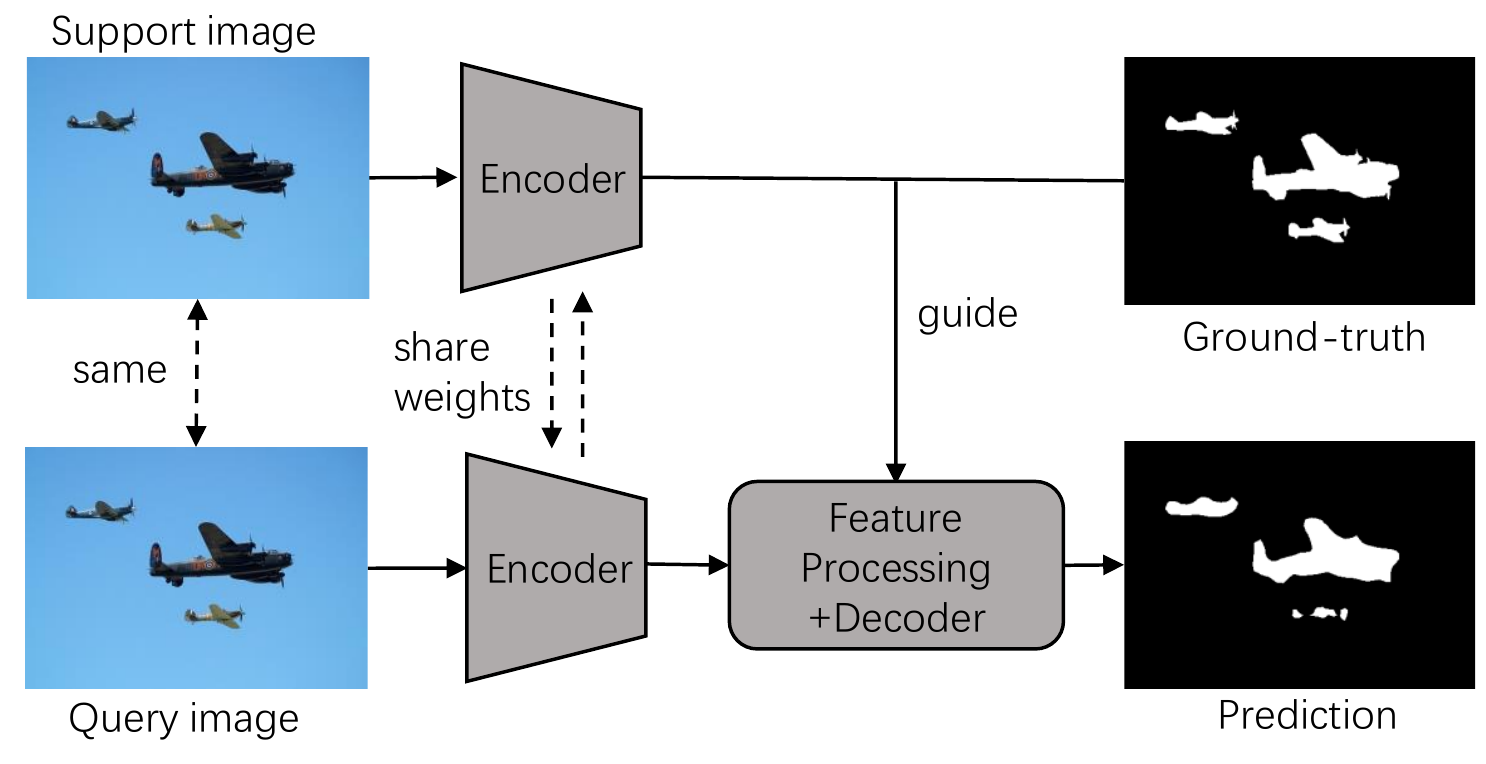}
	\caption{Motivation of our approach. Even using the same image as both support and query input, previous approaches cannot generate accurate segmentation under the guide of its ground-truth mask.}
	\label{fig:motivation}
\end{figure}
Most approaches~\cite{dong2018few, liu2020part, wang2019panet, zhang2019canet, yang2020new, tian2020differentiable} adopt a Siamese Convolutional Neural Network (SCNN) to encode both support and query images. In order to apply the information from support images, they mainly use masked Global Average Pooling (GAP)~\cite{zhou2016learning} or other strengthened methods~\cite{nguyen2019feature} to extract all foreground~\cite{wang2019panet, zhang2019canet, liu2020crnet} or background~\cite{wang2019panet} as one feature vector, which is used as a prototype to compute cosine distance~\cite{zhang2020sg} or make dense comparison~\cite{zhang2019canet} on query images. 

Using a support feature vector extracted from the support image does facilitate the query image segmentation, but it does not carry sufficient information. Fig.~\ref{fig:motivation} shows an extreme example where the support image and query image are exactly the same. However, even the existing best performing approaches fail to accurately segment the query image. We argue that when we use masked GAP or other methods~\cite{nguyen2019feature} to encode a support image to a feature vector, it is unavoidable to lose some useful information due to the average operation. Using such a feature vector to guide the segmentation cannot make a precise prediction for pixels which need the lost information as support. Furthermore, for the multiple shot case such as 5-shot segmentation, the common practice is to use the average of predictions from 5 individual support images as the final prediction~\cite{zhang2020sg} or the average of 5 support vectors as the final support vector~\cite{wang2019panet}. However, the quality of different support images is different, using an average operation forces all support images to share the same contribution.

In this paper, we propose a simple yet effective Self-Guided and Cross-Guided Learning approach (SCL) to overcome the above mentioned drawbacks. Specifically, we design a Self-Guided Module (SGM) to extract comprehensive support information from the support set. Through making an initial prediction for the annotated support image with the initial prototype, the covered and uncovered foreground regions are encoded to the primary and auxiliary support vectors using masked GAP, respectively. By aggregating both primary and auxiliary support vectors, better segmentation performances are obtained on query images.

Enlightened by our proposed SGM, we propose a Cross-Guided Module (CGM) for multiple shot segmentation, where we can evaluate prediction quality from each support image using other annotated support images, such that the high-quality support image will contribute more in the final fusion, and vice versa. Compared to other complicated approaches such as the attention mechanism~\cite{zhang2019canet, zhang2019pyramid}, our CGM does not need to re-train the model, and directly applying it during inference can improve the final performance. Extensive experiments show that our approach achieves new state-of-the-art performances on PASCAL-$\text{5}^i$ and COCO-$\text{20}^i$ datasets.

Our contributions are summarized as follows: 

\begin{itemize}	
	
	\item  We observe that it is unavoidable to lose some useful critical information using the average operation to obtain the support vector. To mitigate this issue, we propose a self-guided mechanism to mine more comprehensive support information by reinforcing such easily lost information, thus accurate segmentation mask can be predicted for query images. 
	
	\item We propose a cross-guided module to fuse multiple predictions from different support images for the multiple shot segmentation task. Without re-training the model, it can be directly used during inference to improve the final performance.
	
	\item Our approach can be applied to different baselines to improve their performance directly. Using our approach achieves new state-of-the-art performances on PASCAL-$\text{5}^i$ (mIoU for 1-shot: 61.8\%, 5-shot: 62.9\%) and COCO-$\text{20}^i$ datasets (mIoU for 1-shot: 37.0\%, 5-shot: 39.9\%) for this task. 
\end{itemize}

\section{Related Work}
\subsection{Fully Supervised Semantic Segmentation}
Fully supervised semantic segmentation, requiring to make pixel-level prediction, has been boosted by recent advances in Convolutional Neural Network (CNN) especially FCN~\cite{long2015fully}. Many network frameworks have been designed based on FCN. For example, UNet~\cite{ronneberger2015u} adopted a multi-scale strategy and a convolution-deconvolution architecture to improve the performance of FCN~\cite{long2015fully}, while PSPNet~\cite{zhao2017pyramid} was proposed to use the pyramid pooling module to generate object details. Deeplab~\cite{chen2018deeplab, chen2018encoder} designed an Atrous Spatial Pyramid Pooling (ASPP)~\cite{chen2017rethinking} module and used dilated convolution~\cite{chen2014semantic} to the FCN architecture.  

\subsection{Few-Shot Segmentation}
Most previous approaches adopt a metric learning strategy~\cite{hu2019attention, vinyals2016matching, sung2018learning, azad2020texture, lei2020few} for few-shot segmentation. For example, In PL~\cite{dong2018few}, a two-branch prototypical network was proposed to segment objects using metric learning. SG-One~\cite{zhang2020sg} proposed to compute a cosine similarity between the generated single support vector and query feature maps to guide the segmentation process. CANet~\cite{zhang2019canet} designed a dense comparison module to make comparisons between the support vector and query feature maps. PANet~\cite{wang2019panet} introduced a module to use the predicted query mask to segment the support images, where it still relied on the generated support vector. FWB~\cite{nguyen2019feature} tried to enhance the feature representation of generated support vector using feature weighting while CRNet~\cite{liu2020crnet} focused on utilizing co-occurrent features from both query and support images to improve the prediction, and it still used a support vector to guide the final prediction. PPNet~\cite{liu2020part} tried to generate  prototypes for different parts as support information. PFENet~\cite{tian2020prior} designed a multi-scale module as decoder to utilize the generated single support vector. 

However, most approaches used masked GAP~\cite{zhou2016learning} or some more advanced methods such as FWB~\cite{nguyen2019feature} to fuse all foreground or background features as a single vector, which unavoidably loses some useful information. Our proposed method tries to provide comprehensive support information using a self-guided approach. 

\begin{figure*}
	\centering
	\includegraphics[width=0.95\textwidth]{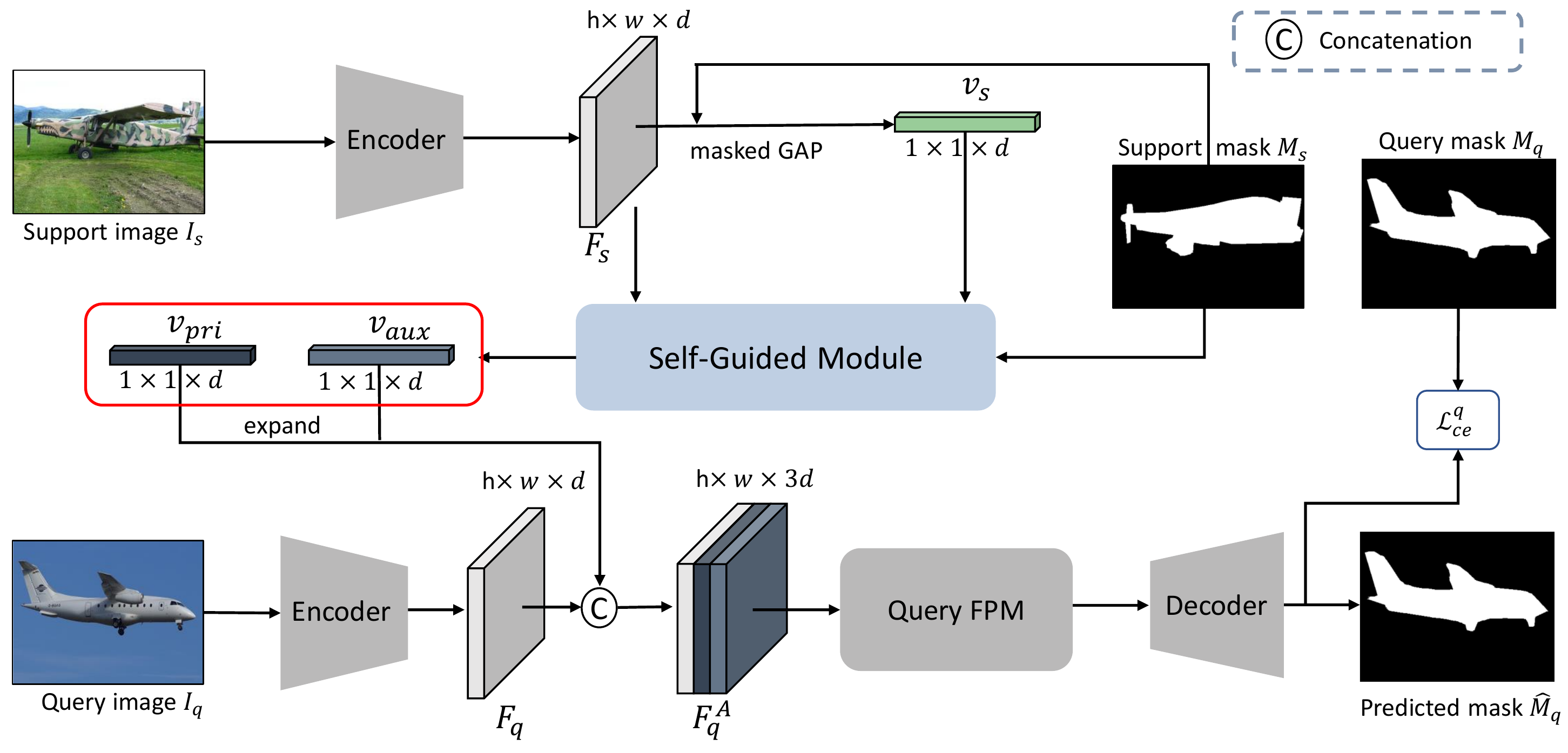}
	\caption{The framework of our SCL approach for 1-shot segmentation. We firstly use an encoder to generate feature maps $F_s$ and $F_q$ from a support image and a query image, respectively. Then masked GAP is used to generate the initial support vector $v_s$. After that, our proposed self-guided module (SGM) takes $v_s$ and $F_s$ as input and output two new support vectors $v_{pri}$ and $v_{aux}$, which are then used as the support information to segment the query image. Encoders for support and query images share the same weights.}
	\label{fig:framework}
\end{figure*}

\section{Problem Setting}
The purpose of few-shot segmentation is to learn a segmentation model which can segment unseen objects provided with a few annotated images of the same class. We need to train a segmentation model on a dataset $D_{\text{train}}$ and evaluate on a dataset $D_{\text{test}}$. Suppose the classes set in $D_{\text{train}}$ is $C_{\text{train}}$ and the classes set in $D_{\text{test}}$ is $C_{\text{test}}$,  there is no overlap between training set and test set, \ie, $C_{\text{train}} \cap C_{\text{test}} = \varnothing$.

Following the previous definition in~\cite{shaban2017one}, episodes are applied to both training set $D_{\text{train}}$ and test set $D_{\text{test}}$ to set a $K$-shot segmentation task. Each episode is composed of a support set $S$ and a query set $Q$ for a specific class $c$. For one episode, the support set contains $K$ images and their masks, \ie, $S = \left\{(I_s^i,M_s^{i})\right\}_{i=1}^{K}$, where $I_s^i$ represents the \emph{i}th image and $M_s^{i}$ indicates its binary mask for the class $c$. A query set contains $N$ images and their binary masks for the class $c$, \ie, $Q = \left\{(I_q^i,M_q^{i})\right\}_{i=1}^{N}$, where $M_q^{i}$ is only used for training. For clear description, we use $S_{\text{train}}$ and $Q_{\text{train}}$ to represent the training support set and query set, while $S_{\text{test}}$ and $Q_{\text{test}}$ for the test set. A model is learned using the training support set $S_{\text{train}}$ and query set $Q_{\text{train}}$. Then the model is evaluated on $D_{\text{test}}$ using the test support set $S_{\text{test}}$ and query set $Q_{\text{test}}$.


\section{Methodology}

\subsection{Overview}
\begin{figure*}
	\centering
	\includegraphics[width=0.95\textwidth]{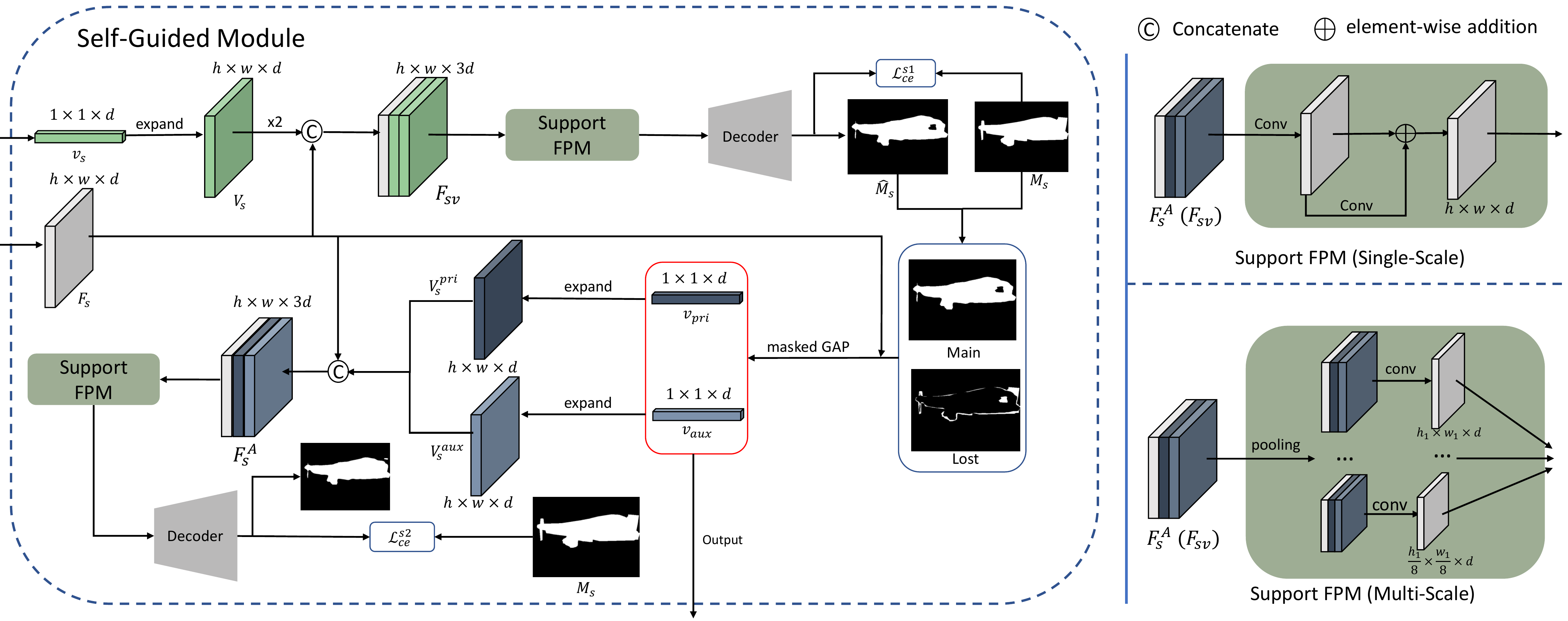}
	\caption{The details of our proposed SGM. Our SGM uses the feature map $F_s$ and the support vector $v_s$ of the support image as input, and produces two new support vectors $v_{pri}$ and $v_{aux}$. In order to provide high-quality support vectors, the support image mask is used as supervision. We provide two kinds of support Feature Processing Modules (FPM) to adapt to different decoders. All support FPMs share the same weights and all decoders are shared with the decoder in Fig~\ref{fig:framework}.}
	\label{fig:SGM}
\end{figure*}

Fig.~\ref{fig:framework} shows our framework for 1-shot segmentation, which can be divided into the following steps:

\begin{enumerate}[1)]
	\item Both support and query images are input to the same encoder to generate their feature maps. After that, an initial support vector is generated using masked GAP from all foreground pixels of the support image. 
	\item With the supervision of the support image mask, our SGM produces two new feature vectors including the primary and auxiliary support vectors,  using the initial support vector and support feature map as input. 
	\item In this step, the primary and auxiliary support vectors are concatenated with the query feature map to guide the segmentation of query images. Through a query Feature Processing Module (FPM) and a decoder, the segmentation mask for the query image is generated. Note that all encoders and decoders are shared. 
\end{enumerate}

\subsection{Self-Guided Learning on Support Set}\label{sec:SGM}
Self-Guided module (SGM) is proposed to provide comprehensive support information to segment the query image. The details of our SGM can be found in Fig.~\ref{fig:SGM}.

Suppose the support image is $I_s$, after passing through the encoder, its feature maps is $F_s$. Then we use masked GAP to generate the initial support vector following previous approaches~\cite{zhang2019canet,zhang2020sg,zhu2020self}: 
\begin{equation}
v_s = \frac{\sum\limits_{i=1}^{hw} F_s(i)\cdot[M_s(i)=1]}{\sum\limits_{i=1}^{hw}[M_s(i)=1]},
\end{equation}
where $i$ is the index of the spatial position. $h$ and $w$ are the height and width of the feature map, respectively. $\left[\cdot \right]$ is Iverson bracket, which equals to 1 if the inside condition is true, otherwise equals to 0. $M_s$ is a binary mask and $M_s(i) =1$ indicates the \emph{i}th pixel belongs to class $c$. Note that $M_s$ needs to be downsampled to the same height and width as $F_s$.  

Both $F_s$ and $v_s$ are input to our proposed self-guided module (SGM). The initial feature vector $v_s$ is firstly duplicated and expanded to the same size with $F_s$ following \cite{tian2020prior,zhang2019canet}, represented as $V_s$, which is then concatenated with $F_s$ to generate a new feature map:
\begin{equation}
F_{sv} = \text{\emph{Concat}}(\left[F_s, V_s, V_s \right]),
\end{equation}
where $\text{\emph{Concat}}(\cdot)$ is the concatenation operator.

Then, the probability map for the support image is generated after passing through the support FPM and the decoder:
\begin{equation}
P_{s1} = \text{\emph{softmax}}(\mathcal{D}(\emph{\text{FPM}}_s(F_{sv}))),
\label{eq:ps1}
\end{equation} 
where $P_{s1}$ is the predicted probability map, \ie, $P_{s1} \in \mathbb{R}^{h \times w \times 2}$. $\mathcal{D}(\cdot)$ means the decoder and details can be found in Sec.~\ref{sec:ID}. \emph{softmax} is the softmax layer. $\emph{\text{FPM}}_s(\cdot)$ is the support FPM, as shown in Fig.~\ref{fig:SGM}. According to the requirements of different decoders, we design two kinds of support FPMs: one for providing single-scale input to the decoder~\cite{zhang2019canet,yang2020new} and the other one for providing multi-scale input to the decoder~\cite{tian2020prior}. 

Then the predicted mask is generated from $P_{s1}$:
\begin{equation}
\hat{M}_{s} = \argmax(P_{s1}), 
\end{equation}
where $\hat{M}_{s}$ is a binary mask, in which element 0 is the background and 1 is the indicator for being class $c$.

Using the predicted mask $\hat{M}_{s}$ and the ground-truth mask $M_{s}$, we can generate the primary support vector $v_{pri}$ and the auxiliary support vector $v_{aux}$:
\begin{equation}
v_{pri} = \frac{\sum\limits_{i=1}^{hw} F_s(i)\cdot[M_s(i)=1] \cdot[\hat{M}_s(i)=1]}{\sum\limits_{i=1}^{hw}[M_s(i)=1]\cdot[\hat{M}_s(i)=1]},\label{eq:v_m}
\end{equation}
\begin{equation}
v_{aux} = \frac{\sum\limits_{i=1}^{hw} F_s(i)\cdot[M_s(i)=1] \cdot[\hat{M}_s(i)\neq 1]}{\sum\limits_{i=1}^{hw}[M_s(i)=1]\cdot[\hat{M}_s(i)\neq 1]}.\label{eq:v_l}
\end{equation}

\begin{figure}
	\centering
	\includegraphics[width=0.95\columnwidth]{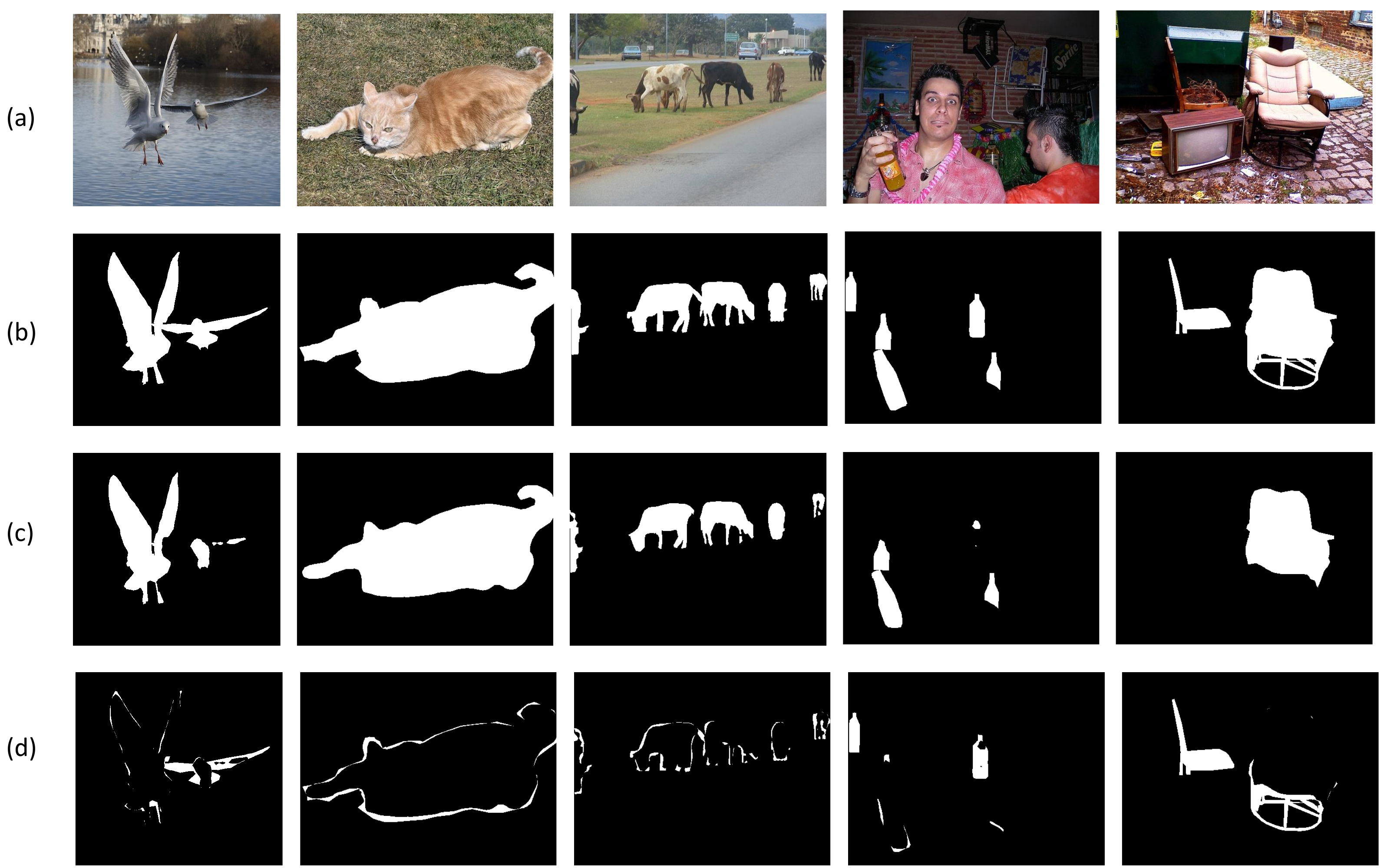}
	\caption{Visualization of the masks for generating $v_{pri}$ and $v_{aux}$. (a) original images. (b) ground-truth (masks for generating $v_s$). (c) masks for generating $v_{pri}$. (d) masks for generating $v_{aux}$. In most cases, $v_{pri}$ aggregates the main information of the support image and $v_{aux}$ mainly collects edge information. In some special cases (the last two columns), $v_{pri}$ loses some body information and $v_{aux}$ encodes all the lost information.}
	\label{fig:v1v2vis}
\end{figure}

In Eq.~(\ref{eq:v_m}), $[M_s(i)=1] \cdot[\hat{M}_s(i)=1]$ indicates the correctly predicted foreground mask using the initial support vector $v_s$ as support. In Eq.~(\ref{eq:v_l}), $[M_s(i)=1] \cdot[\hat{M}_s(i)\neq 1]$ indicates the missing foreground mask. From Eq.~(\ref{eq:v_m}) and Eq.~(\ref{eq:v_l}), it can be found that $v_{pri}$ keeps the main support information as it focuses on aggregating correctly predicted information, $v_{aux}$ focuses on collecting the lost critical information which cannot be predicted using $v_s$. Fig.~\ref{fig:v1v2vis} shows more examples about the masks to produce $v_{pri}$ and $v_{aux}$. It can be seen that $v_{pri}$ ignores some useful information unavoidably while $v_{aux}$ collect all the lost information in $v_{pri}$.

In order to guarantee $v_{pri}$ can collect most information from the support feature map, a cross-entropy loss is used on $P_{s1}$ predicted in Eq.~(\ref{eq:ps1}) :
\begin{equation}
\mathcal{L}_{ce}^{s1} = -\frac{1}{hw}\sum\limits_{i=1}^{hw}\sum\limits_{c_j \in {\left\{ 0,1 \right\} }}[M_s(i)=c_j]log(P_{s1}^{c_j}(i)), \label{eq:lce_s}
\end{equation}
where $0$ is the background class and $1$ is the indicator for a specific foreground class $c$. $P_{s1}^{c_j}(i)$ denotes the predicted probability belonging to class $c_j$ for pixel $i$. 

Then we duplicate and expand $v_{pri}$ and $v_{aux}$ to the same height and width with $F_s$, represented as $V_{pri}^s$ and $V_{aux}^s$, respectively. Following previous process, $F_s$, $V_{s}^{pri}$ and $V_{s}^{aux}$ are concatenated to generate a new feature map $F_{s}^{A}$:
\begin{equation}
F_{s}^{A} = \text{\emph{Concat}}(\left[F_s, V_{s}^{pri}, V_{s}^{aux} \right]).
\end{equation}

After that, the predicted probability map $P_{s2}$ is generated based on the new feature map $F_{s}^{A}$: 
\begin{equation}
P_{s2} = \text{\emph{softmax}}(\mathcal{D}(\text{\emph{FPM}}_s(F_{s}^{A}))). 
\end{equation} 

Similar with Eq.~(\ref{eq:lce_s}), we use a cross-entropy loss to ensure aggregating $v_{pri}$ and $v_{aux}$ together can produce accurate segmentation mask on the support image:
\begin{equation}
\mathcal{L}_{ce}^{s2} = -\frac{1}{hw}\sum\limits_{i=1}^{hw}\sum\limits_{c_j \in {\left\{ 0,1 \right\} }}[M_s(i)=c_j]log(P_{s2}^{c_j}(i)).\label{eq:lce_ml}
\end{equation}


We only use foreground pixels to produce support vectors since background is more complicated than the foreground. Therefore, we cannot guarantee the support vector from background is far away from that of the foreground. 
  
\subsection{Training on Query Set}
Using our proposed SGM, we generate the primary support vector $v_{pri}$ and auxiliary support vector $v_{aux}$, where $v_{pri}$ contains the primary information of support image and $v_{aux}$ collects the lost information in $v_{pri}$. 

Using the same encoder with $I_s$, we also generate the query feature map $F_q$, then $v_{pri}$ and $v_{aux}$ are duplicated and expanded to the same height and width as $F_q$, both of which are then concatenated with $F_q$ to generate a new feature map:
\begin{equation}
F_{q}^A = \text{\emph{Concat}}(\left[F_q, V_q^{pri}, V_q^{aux} \right]),
\end{equation}
where $F_q$ is the feature map of query image $I_q$, which is generated using the same encoder with the support image $I_s$. $V_q^{pri}$ and $V_q^{aux}$ correspond to expanded results of $v_{pri}$ and $v_{aux}$, respectively.

Then $F_{q}^{A}$ is input to a query FPM followed by a decoder to obtain the final prediction:
\begin{equation}
P_q = \text{\emph{softmax}}(\mathcal{D}(\emph{\text{FPM}}_q(F_q^A))),
\end{equation} 
where $\emph{\text{FPM}}_q(\cdot)$ is the query FPM. $P_q$ is the predicted probability map. (More details about the query FPM and decoder can be found in Sec.~\ref{sec:ID} and our supplement material.)

We use a cross-entropy loss to supervise the segmentation of the query image:
\begin{equation}
\mathcal{L}_{ce}^q = -\frac{1}{hw}\sum\limits_{i=1}^{hw}\sum\limits_{c_j \in {\left\{ 0,1 \right\} }}[M_q(i)=c_j]log(P_q^{c_j}(i)),\label{eq:lce_q}
\end{equation}
where $P_q^{c_j}(i)$ denotes the predicted probability belonging to class $c_j$ for pixel $i$.  

The overall training loss is defined as: 
\begin{equation}
\mathcal{L} = \mathcal{L}_{ce}^{s1}+\mathcal{L}_{ce}^{s2}+\mathcal{L}_{ce}^{q}, 
\end{equation}
where $\mathcal{L}_{ce}^{s1}$, $\mathcal{L}_{ce}^{s2}$ are the loss functions defined by Eq.(\ref{eq:lce_s}) and Eq.(\ref{eq:lce_ml}) in Sec.~\ref{sec:SGM}. 

\subsection{Cross-Guided Multiple Shot Learning }
Enlightened by our SGM for 1-shot segmentation, we extend it to Cross-Guided Module (CGM) for the $K$-shot ($K > 1$) segmentation task. Among the $K$ support images, each annotated support image can guide the query image segmentation individually. Based on this principle, we design our CGM where the final mask is fused using predictions from multiple annotated samples with high-quality support images contributing more and vice versa. 

For $K$-shot segmentation task, there are $K$ support images in one episode, \ie, the support set $S = \left\{(I_{s}^1, M_{s}^1), (I_{s}^2, M_{s}^2),..., (I_{s}^K, M_{s}^K) \right\}$. For the \emph{k}th support image $I_{s}^k$, we can firstly use it as the support image and all $K$ support images as query images to input to our proposed 1-shot segmentation model $\mathcal{G}$. The predicted mask for the \emph{i}th support image $I_s^i$ is:
\begin{equation}
\hat{M}_s^{i|k} = \argmax(\mathcal{G}(I_{s}^i | I_{s}^k)),\label{eq:M_ki}
\end{equation} 
where $\hat{M}_s^{i|k}$ is the predicted mask of $I_{s}^i$ under the support of $I_{s}^k$. $\mathcal{G}(I_{s}^i | I_{s}^k)$ outputs the predicted score map of $I_{s}^i$ using $I_{s}^k$ as the support image and $I_{s}^i$ as the query image. 

The ground-truth mask $M_s^i$ for image $I_{s}^i$ is available. Thus, we can evaluate the confident score of $I_{s}^k$ based on the IOU between the predicted masks and their ground-truth masks: 
\begin{equation}
U_s^k = \frac{1}{K}\sum_{i=1}^{K}\text{IOU}(\hat{M}_s^{i|k},M_s^i),\label{eq:U_k}
\end{equation} 
where IOU$(\cdot,\cdot)$ is used to compute the intersection over union score. Then the final predicted score map for an given query image $I_q$ is:
\begin{equation}
\hat{P}_{q} = \text{\emph{softmax}}(\frac{1}{K}\sum_{k=1}^{K}U_s^k\mathcal{G}(I_q|I_{s}^k)).
\label{eq:P_q}
\end{equation} 

A support image with a larger $U_s^k$ makes more contribution to the final prediction, and the generated support vector is more likely to provide sufficient information to segment query images, and vice versa. 

Using CGM does not need to re-train a new model, and we can directly use the segmentation model from 1-shot task to make predictions. Thus, CGM can improve the performance during inference without re-training. 

\section{Experiments}
\begin{table*}[]
	\centering
	\caption{Comparison with other state-of-the-arts using mIoU (\%) as evaluation metric on Pascal-$5^i$ for 1-shot and 5-shot segmentation. ``P." means Pascal. ``\emph{ours-SCL} (CANet)'' and ``\emph{ours-SCL} (PFENet)'' means CANet~\cite{zhang2019canet} and PFENet~\cite{tian2020prior} are applied as baselines, respectively.}\label{tab:VOC_IOU}
	\begin{threeparttable}
	\begin{tabular}{lcccccc|ccccc}
		\hline
		\multirow{2}{*}{Method} & \multirow{2}{*}{Backbone}& \multicolumn{5}{c|}{1-shot}                          & \multicolumn{5}{c}{5-shot}                           \\ \cline{3-12} 
		&& P.-$5^0$ & P.-$5^1$ & P.-$5^2$ & P.-$5^3$ & Mean & P.-$5^0$ & P.-$5^1$& P.-$5^2$ & P.-$5^3$ & Mean \\ \hline
		OSLSM (BMVC'17)~\cite{shaban2017one}&vgg16& 33.6& 55.3& 40.9& 33.5&40.8&35.9&58.1&42.7&39.1&44.0\\
		SG-One~\cite{zhang2020sg}&vgg16& 40.2& 58.4& 48.4& 38.4& 46.3&41.9&58.6&48.6&39.4&47.1 \\
		PANet (ICCV'19)~\cite{wang2019panet}&vgg16&42.3&58.0& 51.1&41.2&48.1&51.8&64.6&59.8&46.5&55.7\\
		PGNet (ICCV'19)~\cite{zhang2019pyramid}&resnet50& 56.0& 66.9& 50.6& 50.4&56.0&57.7&68.7&52.9&54.6& 58.5\\
		CRNet (CVPR'20)~\cite{liu2020crnet}&resnet50& - & - & - & - & 55.7 & -  & -  & - & -  & 58.8 \\
		RPMMs (ECCV'20)~\cite{yang2020prototype}&resnet50&55.2&65.9&52.6&50.7&56.3&56.3&67.3&54.5&51.0&57.3\\
		FWB (ICCV'19)~\cite{nguyen2019feature}&resnet101&51.3& 64.5& 56.7&52.2&56.2&54.8&67.4&62.2&55.3&59.9\\
		PPNet\tnote{*} (ECCV'20)~\cite{liu2020part}&resnet50&47.8&58.8& 53.8&45.6&51.5&58.4&67.8&\textbf{64.9}&56.7&62.0\\ 
		DAN (ECCV'20)~\cite{wang2020few}&resnet101&54.7&68.6&\textbf{57.8}&51.6&58.2&57.9&69.0&60.1&54.9&60.5\\ \hline
		CANet (CVPR'19)~\cite{zhang2019canet}&resnet50& 52.5&65.9&51.3&51.9&55.4&55.5&67.8&51.9&53.2&57.1\\ 
		PFENet (TPAMI'20)~\cite{tian2020prior}&resnet50 &61.7&69.5&55.4&56.3&60.8&63.1&70.7&55.8&57.9&61.9\\ \hline
		\emph{ours-SCL }(CANet) &resnet50&56.8&67.3&53.5&52.5&57.5 &59.5&68.5&54.9&53.7& 59.2\\ 
		\emph{ours-SCL }(PFENet)&resnet50 &\textbf{63.0}&\textbf{70.0}&56.5&\textbf{57.7}&\textbf{61.8} &\textbf{64.5}&\textbf{70.9}&57.3&\textbf{58.7}& \textbf{62.9}\\ \hline
	\end{tabular}
	\begin{tablenotes}
		\item[*] We report the performance without extra unlabeled support data.
	\end{tablenotes}
\end{threeparttable}
\end{table*}
\subsection{Implementation Details}~\label{sec:ID}
Our SCL approach can be easily integrated into many existing few-shot segmentation approaches, and the effectiveness of our approach is evaluated using two baselines: CANet~\cite{zhang2019canet} and PFENet~\cite{tian2020prior}, both of which use masked GAP to generate one support vector for a support image. All decoders in our SGM share the same weights with the decoder in the baseline.

\begin{figure}
	\centering
	\includegraphics[width=\columnwidth]{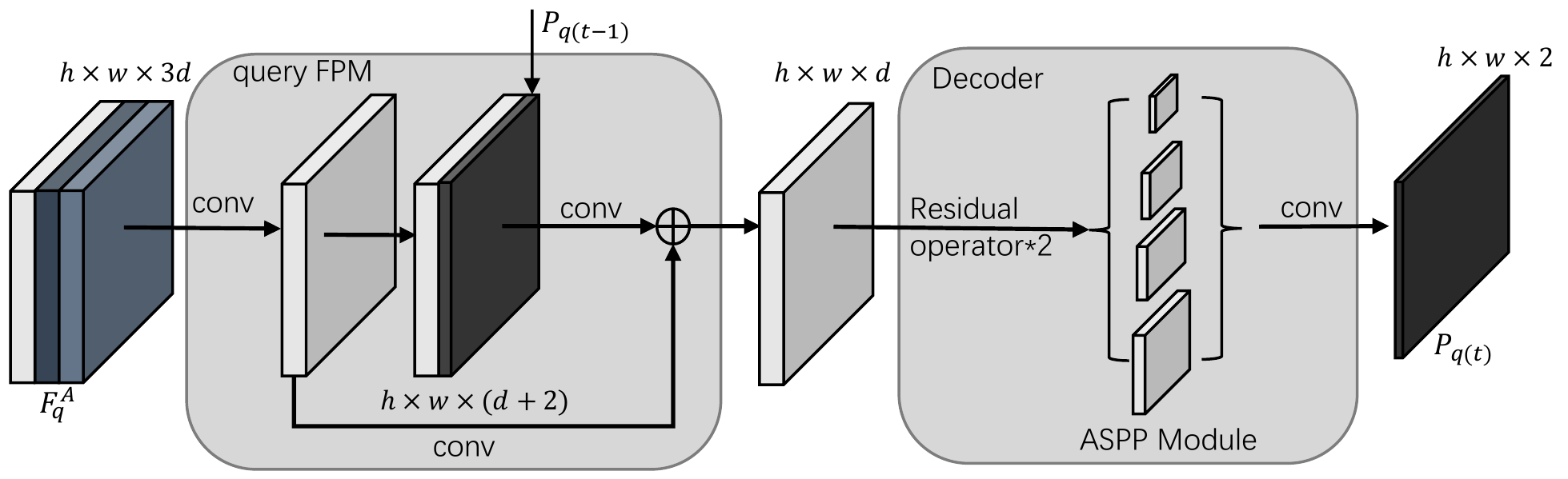}
	\caption{Architecture of the query FPM and decoder in CANet~\cite{zhang2019canet}. CANet used the predicted probability map $P_{q(t-1)}$ from the previous iteration in its query FPM, and its decoder adopts single-scale residual layers following an ASPP module~\cite{chen2017rethinking}.}
	\label{fig:FPM}
\end{figure}

We use single-scale support FPM in our SGM when using CANet~\cite{zhang2019canet} as the baseline since its decoder adopted  single-scale architecture. Besides, the query FPM in CANet~\cite{zhang2019canet} used the probability map $P_{q(t-1)}$ from the previous iteration in the cache to refine the prediction. Fig.~\ref{fig:FPM} shows details of the query FPM and decoder in CANet~\cite{zhang2019canet}.

We use multi-scale support FPM in our SGM when using PFENet~\cite{tian2020prior} as the baseline since its decoder adopted a multi-scale architecture. Additionally, the query FPM in PFENet~\cite{tian2020prior} used a prior mask from the pre-trained model on ImageNet~\cite{russakovsky2015imagenet} as extra support. More details can be found in our supplement material. Note that none of $P_{q(t-1)}$ or the prior mask is used in the support FPM in our SGM.

All training settings are the same as that in CANet~\cite{zhang2019canet} or PFENet~\cite{tian2020prior}. The channel size $d$ in Fig.~\ref{fig:framework} and Fig.~\ref{fig:SGM} is set to 256. The batch size is 4 with 200 epochs used.  The learning rate is 2.5$\times\text{10}^\text{-4}$ and weight decay is 5$\times\text{10}^\text{-4}$ if CANet~\cite{zhang2019canet} is the baseline. The learning rate is 2.5$\times\text{10}^\text{-3}$ and weight decay is 1$\times\text{10}^\text{-4}$ if PFENet~\cite{tian2020prior} is the baseline.

During inference for the 1-shot task, we follow the same settings as in CANet~\cite{zhang2019canet} or PFENet~\cite{tian2020prior}.  
For 5-shot segmentation, we directly use the segmentation model trained on 1-shot task. Following~\cite{wang2019panet}, we average the results from 5 runs with different random seeds as the final performance. All experiments are run on Nvidia RTX 2080Ti.

\subsection{Dataset and Evaluation Metric}
We evaluate our approach on PASCAL-$5^i$ and COCO-$20^i$ dataset. PASCAL-$5^i$ is proposed in OSLSM~\cite{shaban2017one}, which is built based on PASCAL VOC 2012~\cite{everingham2010pascal} and SBD dataset~\cite{hariharan2014simultaneous}. COCO-$20^i$ is proposed in FWB~\cite{nguyen2019feature}, which is built based on MS-COCO~\cite{lin2014microsoft} dataset. 

In PASCAL-$5^i$, 20 classes are divided into 4 splits, in which 3 splits for training and 1 for evaluation. During evaluation, 1000 support-query pairs are randomly sampled from the evaluation set. For more details, please refer to OSLSM~\cite{shaban2017one}. In COCO-$20^i$ , the only difference with PASCAL-$5^i$ is that it divides 80 classes to 4 splits. For more details, please refer to FWB~\cite{nguyen2019feature}. For PASCAL-$5^i$, we evaluate our approach using both CANet~\cite{zhang2019canet} and PFENet~\cite{tian2020prior} as baselines. For COCO-$20^i$, we evaluate our approach based on PFENet~\cite{tian2020prior}.

Following~\cite{wang2019panet}, mean intersection-over-union (mIoU) and foreground-background intersection-over-union (FB-IoU) are used as evaluation metrics.

\subsection{Comparisons with State-of-the-art}
\begin{table*}[]
	\centering
	\caption{Comparison with other state-of-the-arts using mIoU (\%) as evaluation metric on COCO-$20^i$ for 1-shot and 5-shot segmentation. ``C." means COCO-20. ``\emph{ours-SCL} (PFENet)'' means PFENet~\cite{tian2020prior} is applied as the baseline.}\label{tab:COCO}
	\begin{tabular}{lcccccc|ccccc}
		\hline
		\multirow{2}{*}{Method} & \multirow{2}{*}{Backbone}& \multicolumn{5}{c|}{1-shot}                          & \multicolumn{5}{c}{5-shot}                           \\ \cline{3-12} 
		&& C.$^0$ & C.$^1$ & C.$^2$ & C.$^3$ & Mean & C.$^0$ & C.$^1$& C.$^2$ & C.$^3$ & Mean \\ \hline
		FWB (ICCV'19)~\cite{nguyen2019feature}&resnet101&19.9&18.0& 21.0&28.9&21.2&19.1&21.5&23.9&30.1&23.7\\
		PPNet (ECCV'20)~\cite{liu2020part}&resnet50&28.1&30.8&29.5&27.7&29.0&\textbf{39.0}&\textbf{40.8}&37.1&37.3&38.5\\
		DAN (ECCV'20)~\cite{wang2020few}&resnet101&-&-&-&-&24.4&-&-&-&-&29.6\\
		PFENet (TPAMI'20)~\cite{tian2020prior}&resnet101 &34.3&33.0&32.3&30.1&32.4&38.5&38.6&38.2&34.3&37.4\\ \hline
		\emph{ours-SCL }(PFENet)&resnet101 &\textbf{36.4}&\textbf{38.6}&\textbf{37.5}&\textbf{35.4}&\textbf{37.0} &38.9&40.5&\textbf{41.5}&\textbf{38.7}& \textbf{39.9}\\ \hline
	\end{tabular}
\end{table*}

\begin{table}[]
	\centering
	\caption{Comparison with other state-of-the-arts using FB-IoU (\%) on Pascal-$5^i$ for 1-shot and 5-shot segmentation. }\label{tab:VOC_FBIOU}\
	\begin{threeparttable}
	\begin{tabular}{lccc}
		\hline
		\multirow{2}{*}{Method} &\multirow{2}{*}{Backbone}& \multicolumn{2}{c}{FB-IoU (\%)} \\ \cline{3-4} 
		&& 1-shot         & 5-shot         \\ \hline
		CANet (CVPR'19)~\cite{zhang2019canet}&resnet50& 66.2 & 69.6\\
		PFENet (TPAMI'20)~\cite{tian2020prior} \tnote{*}&resnet50& 71.4 & - \\ \hline
		\emph{ours-SCL }(CANet) &resnet50& 70.3&70.7\\
		\emph{ours-SCL }(PFENet) &resnet50& \textbf{71.9}&\textbf{72.8}\\ \hline
	\end{tabular}
	\begin{tablenotes}
		\item[*] The result is generated using models provided by the author.
	\end{tablenotes}
	\end{threeparttable}
\end{table}

In Table~\ref{tab:VOC_IOU}, we compare our approach with other state-of-the-art approaches on PASCAL-$5^i$. It can be seen that our approach achieves new state-of-the-art performances on both 1-shot and 5-shot tasks. Additionally, our approach significantly improves the performances of two baselines on 1-shot segmentation task, with mIoU increases of 2.1\% and 1.0\% for CANet~\cite{zhang2019canet} and PFENet~\cite{tian2020prior}, respectively. For the 5-shot segmentation task, our approach achieves 59.2\% and 62.9\% mIoU using CANet~\cite{zhang2019canet} and PFENet~\cite{tian2020prior}, respectively, both of which are direct improvement without re-training the model. 

In Table~\ref{tab:COCO}, we compare our approach with others on the COCO-$20^i$ dataset. Our approach outperforms other approaches by a large margin, with mIoU gain of 4.6\% and 1.4\% for 1-shot and 5-shot tasks, respectively. 

Table~\ref{tab:VOC_FBIOU} shows the comparison between our approach and two baselines using FB-IoU on PASCAL-$5^i$. Our approach using PFENet~\cite{tian2020prior} as the baseline achieves new state-of-the-art performance. Besides, adopting our approach on CANet~\cite{zhang2019canet} obtain 4.1\% and 1.1\% FB-IoU increases for 1-shot and 5-shot tasks, respectively. 

\begin{figure*}
	\centering
	\includegraphics[width=0.95\textwidth]{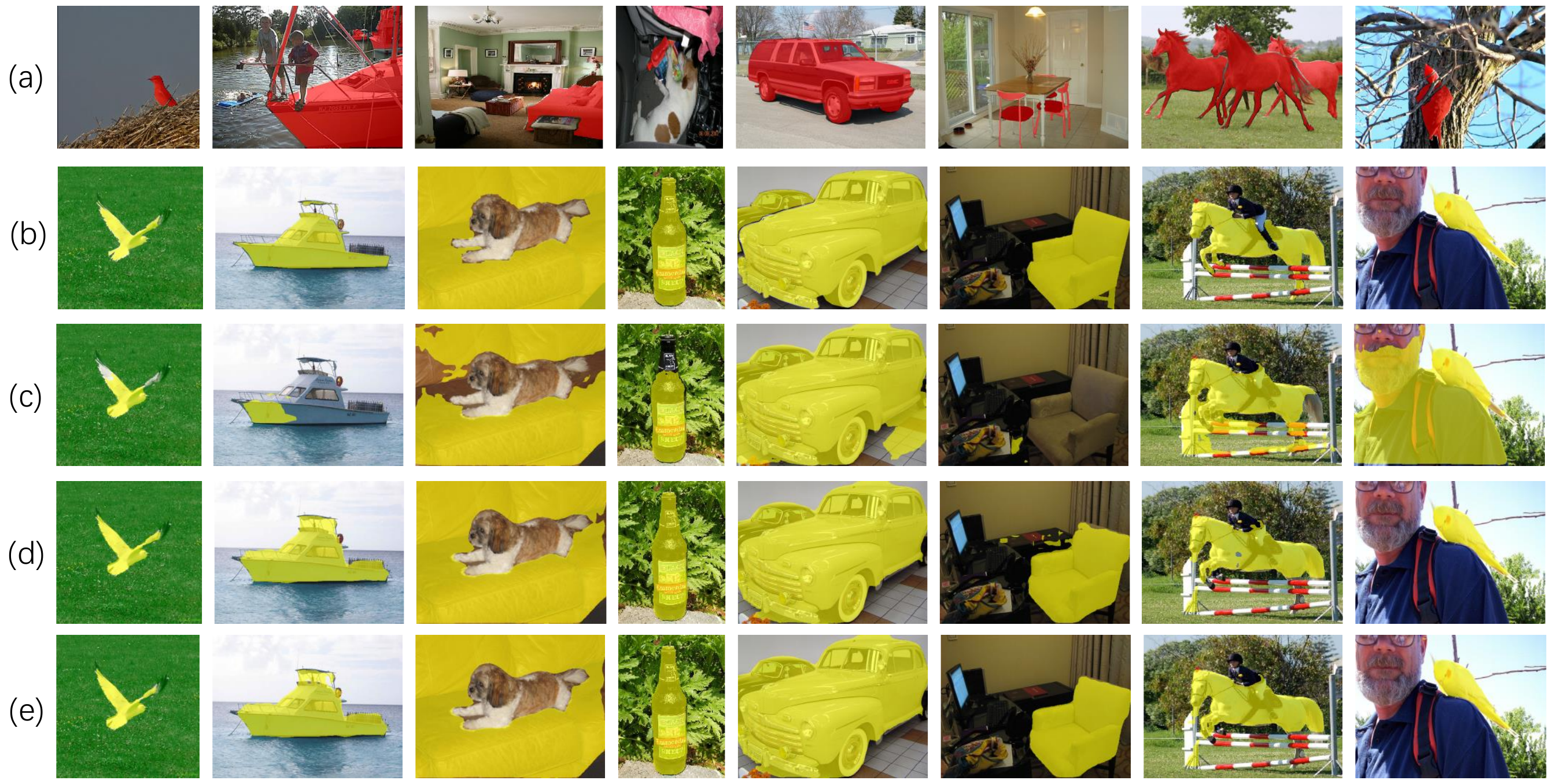}
	\caption{Qualitative results of our approach on Pascal-5$^i$. (a) Support images for the 1-shot task and their masks. (b) Query images and their ground-truth. (c) PFENet~\cite{tian2020prior} 1-shot results. (d) \emph{Ours-SCL} (PFENet) 1-shot results. (e) \emph{Ours-SCL} (PFENet) 5-shot results.}
	\label{fig:vis}
\end{figure*}

In Fig.~\ref{fig:vis}, we report some qualitative results generated by our approach using PFENet~\cite{tian2020prior} as the baseline. It can be seen that our approach produces integral segmentation masks covering object details. More experimental and qualitative results can be found in our supplement material.

\subsection{Ablation Study}

\begin{table}[t]
	\centering
	\caption{Ablation study of our proposed SGM and CGM on PASCAL-$5^i$ for both 1-shot and 5-shot segmentation. ``Avg." means we use the average score of predictions from multiple support images. ``base." means the baseline, which only uses the initial support vector without $\mathcal{L}_{ce}^{s1}$.}\label{tab:CGM}
	\resizebox{0.9\columnwidth}{!}{$
	\begin{tabular}{ccccccc}
		\hline
		shot&base. & SGM & Avg. & CGM & mIoU & FB-IoU\\ \hline
		1&\checkmark&     &-&-&55.4&66.2\\
		1&\checkmark&\checkmark&-&-&\textbf{57.5}&\textbf{70.3}\\ \hline
		5&\checkmark&     &\checkmark&&55.9&66.7\\
		5&\checkmark&&&\checkmark &56.9&69.7\\
		5&\checkmark&\checkmark&\checkmark&& 58.7 & 70.3 \\
		5&\checkmark&\checkmark&         &\checkmark&\textbf{59.2}&\textbf{70.7}\\ \hline
	\end{tabular}
	$}
\end{table}

In this section, we conduct ablation studies on PASCAL-$5^i$ using CANet~\cite{zhang2019canet} as the baseline and all results are average mIoU across 4 splits. 

We firstly conduct an ablation study to show the influence of our proposed SGM and CGM in Table~\ref{tab:CGM}. For 1-shot, compared with the baseline, using SGM improves the performance by a large margin, being 2.1\%  and 4.1\% for mIoU and FB-IoU, respectively. For 5-shot, using both SGM and CGM together obtains a 59.2\% mIoU score, which is 3.3\% higher compared to the baseline with the average method. Compared with the average method, our CGM directly increases the mIoU score by 0.5\% when SGM is adopted. It is worth to notice that our CGM does not need to re-train the model and the gain is obtained in the inference stage. 

\begin{table}[t]
	\centering
	\caption{Ablation study of the support vectors in our proposed SGM on PASCAL-$5^i$ for 1-shot segmentation. $v_s$, $v_{pri}$ and $v_{aux}$ are initial, primary and auxiliary feature vectors generated by our SGM, respectively. Note that $\mathcal{L}_{ce}^{s1}$ is used for $v_s$.} \label{tab:SMM} 
	\begin{tabular}{ccccc}
		\hline
		$v_s$& $v_{pri}$ & $v_{aux}$ & mIoU (\%) &FB-IoU (\%) \\ \hline
		\checkmark &    &     &55.6&67.3\\
		& \checkmark &     &56.6&69.5\\
		&    & \checkmark  &51.4&65.2\\
		\checkmark & \checkmark & \checkmark  &57.1&69.9\\
		&\checkmark & \checkmark  &\textbf{57.5}&\textbf{70.3} \\ \hline
	\end{tabular}
	
\end{table}

\begin{table}[t]
	\centering
	\caption{Ablation study of loss functions in the SGM on PASCAL-$5^i$ for 1-shot segmentation. $\mathcal{L}_{ce}^{s1}$ means the loss function in Eq.~(\ref{eq:lce_s}). $\mathcal{L}_{ce}^{s2}$ means the loss function in Eq.~(\ref{eq:lce_ml}).}\label{tab:SMM_loss}
	\begin{tabular}{cccc}
		\hline
		$\mathcal{L}_{ce}^{s1}$ & $\mathcal{L}_{ce}^{s2}$ & mIoU (\%) &FB-IoU (\%) \\ \hline
		\checkmark &    & 55.6&67.3\\
		& \checkmark &56.8&69.6\\
		\checkmark & \checkmark &\textbf{57.5}&\textbf{70.3} \\ \hline
	\end{tabular}
\end{table}

Table~\ref{tab:SMM} shows the influence of the support vectors on the proposed SGM for 1-shot segmentation. If only $v_s$ is adopted, the mIoU and FB-IoU scores are 55.6\% and 67.3\% respectively. Using SGM (with both $v_{pri}$ and $v_{aux}$) achieves 57.5\% and 70.3\% on mIoU and FB-IoU, with a significant gain of 1.9\% and 3.0\% on mIoU and FB-IoU, respectively. Besides, It can also be seen that when using $v_{pri}$ and $v_{aux}$ individually, it only achieves 56.6\% and 51.4\% on mIoU, both of which are much lower than using them jointly. Solely using $v_{aux}$ even performs worse than the baseline (only using $v_s$). Furthermore, we also evaluate the performance when using all support vectors ($v_s$, $v_{pri}$ and $v_{aux}$) together, it can be seen that it does not improve the results, which also proves that $v_{pri}$ and $v_{aux}$ already provide sufficient information as support, demonstrating the effectiveness of our SGM. Note that when using all support vectors, channels of $F_q^A$ should be increased to $4d$. 

Table~\ref{tab:SMM_loss} studies the influence of loss functions $\mathcal{L}_{ce}^{s1}$ and $\mathcal{L}_{ce}^{s2}$ in SGM. Using both $\mathcal{L}_{ce}^{s1}$ and $\mathcal{L}_{ce}^{s2}$ significantly outperforms the baseline. If only $\mathcal{L}_{ce}^{s1}$ is adopted without $\mathcal{L}_{ce}^{s2}$, the obtained mIoU score is 55.6\%, being 1.9\% lower than using both loss functions together. This is because  $\mathcal{L}_{ce}^{s2}$ provides one more step of training by treating the support image as query image, where both support vectors $v_{pri}$ and $v_{aux}$ are deployed. Similarly, if only $\mathcal{L}_{ce}^{s2}$ is adopted without $\mathcal{L}_{ce}^{s1}$, the obtained performance is also lower than using both loss functions together. This is because using $\mathcal{L}_{ce}^{s1}$ can ensure primary support vector $v_{pri}$ focus on extracting the main information while $v_{aux}$ focus on the lost information. Without $\mathcal{L}_{ce}^{s1}$, the roles of $v_{pri}$ and $v_{aux}$ get mixed and vague.

\section{Conclusion}
We propose a self-guided learning approach for few-shot segmentation. Our approach enables to extract comprehensive support information using our proposed self-guided module. Besides, in order to improve the drawbacks of average fusion for multiple support images, we propose a new cross-guided module to make highly quality support images contribute more in the final prediction, and vice versa. Extensive experiments show the effectiveness of our proposed modules. In the future, we will try to use the background information as extra support to improve our approach.

{\small
	\bibliographystyle{ieee_fullname}
	\bibliography{ref}
}
\end{document}